\documentclass[10pt,twocolumn,letterpaper]{article}

\usepackage{conf}
\usepackage{times}
\usepackage{inconsolata}
\usepackage[utf8]{inputenc}
\usepackage[T1]{fontenc}

\usepackage{microtype}
\usepackage{graphicx}
\usepackage{subcaption}
\usepackage{booktabs}
\usepackage{tabularx}
\newcolumntype{Y}{>{\centering\arraybackslash}X}

\usepackage[round]{natbib}

\usepackage{algorithm}

\usepackage[table,dvipsnames]{xcolor}
\usepackage[pagebackref=true,breaklinks=true,colorlinks,bookmarks=false]{hyperref}
\definecolor{mydarkblue}{rgb}{0,0.08,0.45}
\hypersetup{
  linkcolor=mydarkblue,
  citecolor=mydarkblue,
  filecolor=mydarkblue,
  urlcolor=mydarkblue,
}

\usepackage{amsmath}
\usepackage{amssymb}
\usepackage{mathtools}
\usepackage{amsthm}

\usepackage[capitalize,noabbrev]{cleveref}

\theoremstyle{plain}

\theoremstyle{definition}

\theoremstyle{remark}

\usepackage[disable,textsize=tiny]{todonotes}

\usepackage{algpseudocode}
\usepackage{xspace}

\newcommand{\flashOptim}{FlashOptim\xspace}
\newcommand{\flashAdam}{FlashAdamW\xspace}
\newcommand{\flashSGD}{FlashSGD\xspace}
\newcommand{\flashLion}{FlashLion\xspace}

\definecolor{ultramarine}{RGB}{0,32,96}

\newcommand{\hlcolor}{SpringGreen}  
\newcommand{\hlchange}[1]{{{\fboxsep=1pt\text{\colorbox{\hlcolor}{$#1$}}}}}
\newcommand{\hltext}[1]{\colorbox{\hlcolor}{#1}}

\definecolor{firebrick}{RGB}{178,34,34}
\definecolor{navy}{RGB}{0,0,128}
\definecolor{forestgreen}{RGB}{0,128,0}

\ifdefined\final
  \newcommand{\authorcomment}[3]{}
  \newcommand{\draft}[1]{}
  
\else
  \newcommand{\authorcomment}[3]{{\color{#2}[\textbf{#1}:#3]}}
  \newcommand{\draft}[1]{{\color{ultramarine}\itshape #1}}
  
\fi

\usepackage{mathtools}

\usepackage{amsmath}

\newcommand{\thetalp}{\theta^\prime}

\algtext*{EndFunction}
\algtext*{EndProcedure}
\algtext*{EndIf}
\algtext*{EndFor}
\algtext*{EndWhile}
\algtext*{EndLoop}
\algtext*{EndRepeat}

\algrenewcommand{\algorithmiccomment}[1]{// #1}

\newcommand{\nseed}{3\xspace}

\usepackage{enumitem}

\usepackage{siunitx}
\usepackage{setspace}

\conffinalcopy

\begin{document}

\title{FlashOptim: Optimizers for Memory-Efficient Training}

\author{Jose Javier Gonzalez Ortiz\thanks{Equal contribution}\\
Databricks AI Research\\
{\tt\small j.gonzalez@databricks.com}
\and
Abhay Gupta\\
Databricks AI Research\\
{\tt\small abhay.gupta@databricks.com}
\and
Christopher Rinard\\
Databricks AI Research\thanks{Now at Standard Kernel Co.}\\
{\tt\small chris@standardkernel.co}
\and
Davis Blalock\footnotemark[1]\\
Databricks AI Research\thanks{Now at Google DeepMind}\\
{\tt\small daviswblalock@gmail.com}
}

\maketitle

\begin{abstract}

Standard mixed-precision training of neural networks requires many bytes of accelerator memory for each model parameter. These bytes reflect not just the parameter itself, but also its gradient and one or more optimizer state variables. With each of these values typically requiring 4 bytes, training even a 7 billion parameter model can be impractical for researchers with less than 100\,GiB of accelerator memory.

We introduce \flashOptim, a suite of optimizations that reduces per-parameter memory by over 50\% while preserving model quality and API compatibility.
Our approach introduces two key techniques. First, we improve master weight splitting by finding and exploiting a tight bound on its quantization error. Second, we design companding functions that greatly reduce the error in 8-bit optimizer state quantization.
Together with 16-bit gradients, these techniques reduce AdamW memory from 16 bytes to 7 bytes per parameter, or 5 bytes with gradient release. They also cut model checkpoint sizes by more than half.

Experiments with \flashOptim applied to SGD, AdamW, and Lion show no measurable quality degradation across a collection of standard vision and language benchmarks, including Llama-3.1-8B finetuning.

\end{abstract}

\section{Introduction} \label{sec:intro}

Recent advances in deep learning have been driven largely by scaling: larger models trained on more data consistently yield better results across language~\citep{kaplan2020scaling,hoffmann2022training,chowdhery2023palm} and vision~\citep{rosenfeld2019constructive,tan2019efficientnet,dehghani2023scalingvit} domains.
Training large models can require a great deal of accelerator memory, with each training iteration requiring memory to store parameters, activations, gradients, and optimizer state.

How much memory do these tensors require? \autoref{tbl:bytesPerParam} shows a typical breakdown. Excluding activations, which scale with batch size rather than parameter count, training with Adam uses about 16 bytes per parameter. Training a 7-billion-parameter LLM therefore requires at least 112\,GiB of accelerator memory, plus more memory for activations.

\begin{table}[t]
    \centering
    \caption{\textbf{Memory per parameter (bytes) for model training}. 
    \flashOptim reduces Adam from 16 to 7 bytes and SGD from 12 to 6 bytes.
     $(\star)$ With gradient release, we further reduce total memory requirements by 2 bytes.}
    \label{tbl:bytesPerParam}
    \setlength{\tabcolsep}{3pt}

    \small
    \rowcolors{2}{white}{gray!10}
    \renewcommand{\arraystretch}{1.2}
    \begin{tabular}{lcccc}
        \toprule
        \textbf{Tensor} & \textbf{SGD} & \textbf{\flashSGD} & \textbf{Adam} & \textbf{FlashAdam} \\
        \midrule
        Master Weights      & 4 & 2\phantom{ (0${}^\star$)} & 4 & 2\phantom{ (0${}^\star$)} \\
        Weight Correction   &   & 1\phantom{ (0${}^\star$)} &   & 1\phantom{ (0${}^\star$)} \\
        Gradients           & 4 & 2 (0${}^\star$) & 4 & 2 (0${}^\star$) \\
        Momentum            & 4 & 1\phantom{ (0${}^\star$)} & 4 & 1\phantom{ (0${}^\star$)} \\
        Variance            &   &   & 4 & 1\phantom{ (0${}^\star$)} \\
        \midrule
        \textbf{Total} & \textbf{12} & \textbf{6 (4${}^\star$)} & \textbf{16} & \textbf{7 (5${}^\star$)} \\
        \bottomrule
    \end{tabular}
\end{table}

Several approaches mitigate this memory consumption. Distributed training with tensor sharding~\citep{rajbhandari2020zero} divides the memory load across multiple accelerators. While this is standard practice in well-resourced organizations, it requires access to multiple accelerators that many practitioners lack.
A second approach is CPU offloading~\citep{ren2021zero}, which moves some tensors to host memory at the cost of added overhead and complexity. Third, parameter-efficient methods~\citep{li2021prefix,hu2022lora} reduce trainable parameters by freezing most weights and training either a small subset of the original weights or a small set of new auxiliary weights, but fundamentally alter the training dynamics~\citep{biderman2024lora}.

\begin{figure}
    \centering
    \includegraphics[width=\columnwidth]{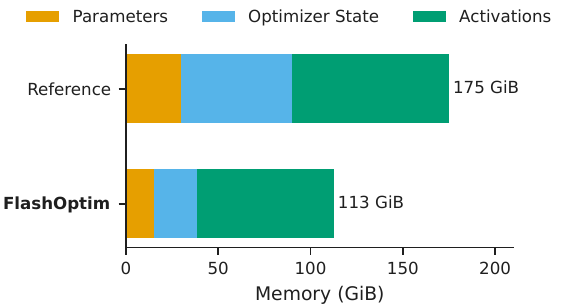}
    \caption{\textbf{Memory breakdown for finetuning Llama-3.1-8B.} \flashOptim reduces peak memory from 175 to 113\,GiB by compressing parameters and optimizer states.}
    \label{fig:finetuning_memory_breakdown}
\end{figure}

In this work, we describe \flashOptim, a set of techniques to reduce 
parameter-associated memory in common deep learning optimizers. 
\autoref{fig:finetuning_memory_breakdown} shows an example: with \flashOptim, finetuning 
Llama-3.1-8B drops from 175\,GiB to 113\,GiB peak memory.
Crucially, these memory savings are effectively free: \flashOptim runs just as fast as standard optimizers and causes no measurable loss of model quality across a suite of established training tasks (\S\ref{sec:results}). This allows our optimizer implementations to serve as drop-in replacements for their unoptimized counterparts. \flashOptim incorporates existing enhancements, such as gradient release~\citep{zhang2023adam,warner2024optimi}, while also introducing improved float splitting~\citep{zamirai2020revisiting,warner2024optimi} and simplified 8-bit optimizer state quantization~\citep{dettmers2022optimizers,peng2023fp8lm,xi2025coat,fishman2024scaling}. Furthermore, \flashOptim composes cleanly with existing memory-reduction techniques, such as sharding tensors across accelerators, offloading to CPU, or freezing parameters.

We make the following contributions:

\begin{itemize}[itemsep=-2pt]
    \item \textbf{Improved float splitting}: Instead of materializing both a 32-bit master weight and a 16-bit downcast weight for forward and backward, one can split each master weight into a low-precision weight and a correction term stored in the optimizer~\citep{zamirai2020revisiting,warner2024optimi}. We improve on existing float splitting techniques by (a) enabling either 8- or 16-bit error correction and (b) achieving much lower reconstruction error for a given number of correction bits. This allows us to use 24-bit master weights with no loss of model quality.
    \item \textbf{Companded optimizer state quantization}: Several works have shown that one can compress optimizer states to 8 bits per element given sufficient software complexity. We demonstrate that one can do this much more simply, with nothing more than a one-line preprocessing function before standard group-wise linear quantization. Our ablations across different tensor types suggest that designing custom companding functions is a fruitful direction for future research. 
    \item \textbf{Fused optimized kernels}: We implement \flashOptim as optimizer step kernels that fuse all compression and quantization operations, reducing memory while preserving throughput during training. Our implementation is publicly available at \url{https://github.com/databricks/flashoptim}.
\end{itemize}

\section{Related Work} \label{sec:relatedWork}

\textbf{Low-Precision Training.}
Mixed-precision training~\citep{micikevicius2018mixed} executes forward and backward passes in FP16 to reduce memory and compute, while retaining FP32 precision for optimizer states and master weights to preserve numerical stability.
\citet{kalamkar2019bfloat16} showed that BFloat16~\citep{google_cloud_bfloat16_2019} works equally well, and \citet{zamirai2020revisiting} explored pure BF16 master weights with stochastic rounding and Kahan summation.
Recent work has pushed further with FP8 training~\citep{wang2018training,mellempudi2019mixed,micikevicius2022fp8,fishman2024scaling,narayan2025mu}, though these approaches primarily target compute formats and retain higher-precision storage for master weights.
\flashOptim extends this line of work with an improved float splitting mechanism that reduces storage to 3 bytes per parameter, down from 4-byte FP32, while maintaining FP32-equivalent training semantics.

\textbf{Optimizer State Compression.}
\citet{dettmers2022optimizers} applied 8-bit block-wise dynamic quantization to Adam's momentum and variance, reducing optimizer state from 8 to 2 bytes per parameter.
Follow-up work explored FP8 representations~\citep{peng2023fp8lm,xi2025coat,fishman2024scaling}, and \citet{li2023memory} compressed both moments to 4 bits using row and column-wise quantization.
MicroAdam~\citep{modoranu2024microadam} instead compresses gradients before updating optimizer states. Rather than design elaborate quantization methods or number formats, we show that one can obtain 8-bit optimizer states with no quality loss using simple, one-line preprocessing functions.
Beyond optimizer states, we address additional sources of per-parameter memory, eliminating entire bytes from other tensors.

\textbf{Gradient Memory and Communication.}
LOMO~\citep{lv2024lomo}, AdaLOMO~\citep{lv2024adalomo}, and Adam Accumulation~\citep{zhang2023adam} fuse parameter updates into the backward pass to release gradient memory eagerly. However, this conflicts with gradient accumulation, which requires the full accumulated gradient before updating.
In distributed settings, gradient communication can also become a bottleneck. One can reduce this bottleneck by, e.g., compressing gradients to 1-bit with error feedback~\citep{tang20211bit}, or using low-rank approximations~\citep{vogels2019powersgd}.
\flashOptim supports gradient release when compatible and could be used alongside communication compression techniques.

\textbf{Memory-Efficient Optimization.}
Alternative optimizer designs reduce memory by restructuring update rules and stored buffers.
Adafactor~\citep{shazeer2018adafactor} achieves sublinear memory by factorizing the second moment into row and column statistics; SM3~\citep{anil2019memory} stores structured maxima; NovoGrad~\citep{ginsburg2019stochastic} replaces per-parameter variance with layer-wise normalization.
Adam-mini~\citep{zhang2025adammini} shares variance terms across parameter blocks, while Adapprox~\citep{zhao2024adapprox} uses a low-rank approximation.
Other approaches eliminate the second moment entirely: Lion~\citep{chen2023lion} uses sign-based momentum, and Muon~\citep{jordan2024muon,liu2025muon} applies orthogonalized updates.
\citet{pethick2025training} extend Muon to unify gradient accumulation with momentum, removing dedicated optimizer memory altogether.

Low-rank decompositions approximate full tensors while requiring less memory. 
For fine-tuning, LoRA~\citep{hu2022lora} and QLoRA~\citep{dettmers2023qlora} freeze base weights and train only low-rank adapters.
For pretraining, GaLore~\citep{zhao2024galore} projects gradients to a low-rank subspace, and APOLLO~\citep{zhu2025apollo} approximates adaptive scaling with random projections.
Unlike these approaches that modify the optimizer's update rule, \flashOptim preserves standard optimizer semantics and can be combined with these techniques.

\textbf{System-Level Memory Optimizations.}
System-level approaches reduce accelerator memory without changing optimization semantics.
Activation checkpointing~\citep{chen2016training,korthikanti2023reducing} trades compute for memory by recomputing activations during the backward pass.
ZeRO~\citep{rajbhandari2020zero} partitions optimizer states, gradients, and parameters across data-parallel ranks, while offloading~\citep{rajbhandari2021zeroinfinity,ren2021zero} moves state to CPU or NVMe memory.
\flashOptim is orthogonal to these approaches: it reduces the per-rank footprint and can be used with ZeRO, FSDP~\citep{zhao2023pytorchfsdp}, and activation checkpointing.

\section{Method} \label{sec:method}

This section describes the two key techniques behind \flashOptim: weight splitting (\S\ref{sub:ulp_weight_compression}) and companded optimizer state quantization (\S\ref{sub:optimizer_state_quantization_with_companding}).
We then describe how to integrate these ideas into common optimizer updates while minimizing associated overhead (\S\ref{sub:optimizer_update}).

\subsection{Weight Splitting}
\label{sub:ulp_weight_compression}

Mixed-precision training uses 16-bit weights for forward and backward passes, but accumulating gradient updates requires higher precision to avoid stagnation~\citep{micikevicius2018mixed}.
Thus, the standard practice is to maintain FP32 precision master weights during training.

However, this introduces waste: the downcast weights take space, but store no information beyond what is saved in the master weights. To eliminate this redundancy, weight splitting~\citep{zamirai2020revisiting,warner2024optimi} instead stores the downcast weights and narrow error-correction values. By combining a 16-bit weight $\thetalp$ with a 16-bit error correction value $\rho$, one has enough information to reconstruct a 32-bit master weight $\theta$ with no redundancy.

The core questions in a weight splitting scheme are 1) how to set $(\thetalp, \rho)$ given $\theta$ and 2) how to estimate $\theta$ given $(\thetalp, \rho)$.  
One obvious approach is to use the high 16 bits of $\theta$ as $\thetalp$ and the low 16 bits as $\rho$. This admits exact reconstruction of $\theta$ for the special case of BF16 $\thetalp$ and FP32 $\theta$, since these formats happen to share the same exponent sizes and offsets. However, this approach does not generalize to other pairs of number formats. It also rounds towards zero instead of towards the nearest low-precision value. 

A more general alternative, used in previous work~\citep{zamirai2020revisiting,warner2024optimi}, is to set $\rho = \theta - \thetalp$, represented as a BF16 value. However, the difference of two floating-point numbers requires as many bits as the wider of the two floats to store exactly,\footnote{Consider, e.g., storing the minimum float32 subnormal, which will be rounded to zero by any narrower datatype.} so a BF16 $\rho$ incurs approximation error. For example, if the rounding error were 1e-5, BF16 could only represent 1.0014e-5. In general, while BF16's wide exponent lets it represent nearly the full range of FP32, its 7 mantissa bits only guarantee a relative error bound of 1/256.  

Our observation is that \textbf{all exponent bits in this scheme are wasted}. 
The exponent of $e \triangleq \theta - \thetalp$ can always be inferred from 
$\thetalp$: under round-to-nearest, $\theta$ must lie within 
$[\thetalp - u/2, \thetalp + u/2]$, where $u = \text{ULP}(\thetalp)$ is the 
unit in the last place~\citep{goldberg1991every}. If $\theta$ were outside 
this interval, it would have rounded to a different value. It therefore 
suffices to encode where $e$ falls within this tiny interval rather than 
across the full FP32 range.

To exploit this observation, we rescale $e$ such that $[-u/2, u/2]$ maps to $[-N, N]$; $N \triangleq 2^b -1$, and then quantize this rescaled $e$ to the nearest $b$-bit integer. That is,
\begin{equation}
\begin{aligned}
    \thetalp &= \text{downcast}(\theta) \\
    \rho &= \text{round}\left( \frac{\theta - \thetalp}{\text{ULP}(\thetalp)/2}\cdot N \right), \\
\end{aligned}
\label{eq:compress}
\end{equation}

To reconstruct $\theta$ from $(\thetalp, \rho)$, we invert this scaling and add the result to $\thetalp$. 
\begin{equation}
\begin{aligned}
\hat{\theta} =    \thetalp + \frac{\rho}{N} \cdot \frac{\text{ULP}(\thetalp)}{2}
\end{aligned}
\label{eq:decompress}
\end{equation}
For tensors of floating-point values, we apply this transformation elementwise.
\autoref{alg:split_reconstruct} provides a lower-level description of the compression and decompression operations with numerical precision considerations.

\begin{algorithm}[h]
    \caption{Weight Splitting}
    \label{alg:split_reconstruct}
    \begin{algorithmic}[1]
    \Statex \textbf{Constants:} $N$ (127 for INT8, 32767 for INT16)
    \Function{$\mathcal{C}$}{$\theta$}
        \State $\thetalp \gets \mathrm{Downcast}(\theta)$
        \State $e \gets \theta - \mathrm{Float32}(\thetalp)$
        \State $\ell \gets \lfloor\mathrm{log_2ULP}(\thetalp)\rfloor - 1$
        \State $h \gets \lfloor -\ell / 2 \rfloor$ \Comment{For numerical stability}
        \State $e_{\mathrm{norm}} \gets (e \cdot 2^h) \cdot 2^{-\ell - h}$
        \State $\rho \gets \mathrm{Int}(\mathrm{Round}(\mathrm{Clamp}(e_{\mathrm{norm}}, -1, 1) \cdot N))$
        \State \Return $\thetalp, \rho$
    \EndFunction
    \vspace{0.25\baselineskip}
    \Function{$\mathcal{C}^{-1}$}{$\thetalp, \rho$} 
        \State $\ell \gets \lfloor\mathrm{log_2ULP}(\thetalp)\rfloor - 1$
        \State $h \gets \lfloor \ell / 2 \rfloor$
        \State $e \gets ((\mathrm{Float32}(\rho) / N) \cdot 2^h) \cdot 2^{\ell - h}$
        \State \Return $\mathrm{Float32}(\thetalp) + e$
    \EndFunction
    \end{algorithmic}
    \end{algorithm}

When using BF16 for $\thetalp$ and INT8 for $\rho$, the compressed representation provides approximately 24 bits of effective precision (16 from BF16 plus 8 from the error term).
This is analogous to the PXR24 format used in high-dynamic-range imaging, which achieves similar precision by rounding 32-bit floats to 24 bits~\citep{kainz2004openexr}.

\subsection{Companded Optimizer State Quantization}
\label{sub:optimizer_state_quantization_with_companding}

For optimizer state variables such as momentum and variance estimates, a common approach is group-wise quantization: dividing tensors into fixed-length groups and mapping values to a lower-precision format like INT8~\citep{dettmers2022optimizers}.
To increase the precision range of the group of values, they are rescaled using the maximum absolute value (\emph{absmax}), which is stored as an additional scale with 32 or 16 bits of precision.
While simple, this \emph{uniform} quantization allocates bins evenly across the value range, implicitly assuming that values are roughly uniformly distributed.
Our measurements show that optimizer state distributions violate this assumption (\S\ref{sec:opt_quant}), and we find that applying nonlinear \emph{companding} functions before quantization can reshape these distributions toward uniformity, improving utilization of quantization bins and reducing quantization error.
As we show in \S\ref{sec:opt_quant}, this companding step is critical: without it, linear quantization of optimizer states causes training to diverge.

We design specialized transformations for each optimizer state type.
For momentum tensors (used in SGD, Adam, and Lion), we first normalize each group by its absmax scale, then apply a softsign-like function:
\begin{equation}
    \phi_m(x) = \frac{2x}{1+|x|} \qquad \phi_m^{-1}(z) = \frac{z}{2-|z|}
\end{equation}
This function compresses extreme values: inputs near $\pm 1$ are pushed toward the center, spreading the momentum distribution more evenly across quantization bins.
In contrast, for variance tensors in Adam, we first apply a square root, then normalize by the group absmax:
\begin{equation}
    \phi_v(x) = \sqrt{x} \qquad \phi_v^{-1}(z) = z^2
\end{equation}
Here the square root is motivated by Adam's variance update
$v_t = \beta_2 v_{t-1} + (1 - \beta_2) g^2$, which accumulates squared gradients, producing heavy-tailed distributions with large dynamic range.
Both transformations satisfy key design criteria: they are exactly invertible, computationally efficient in both directions (one division or square root per element), and require no hyperparameters.

    \begin{algorithm}[t]
        \caption{$\mathcal{Q}_m$: Momentum Quantization}
        \label{alg:quant_mom}
        \begin{algorithmic}[1]
        \Statex \textbf{Constants:} Group size $G = 32$
        \vspace{0.25\baselineskip}
        \Function{$\mathcal{Q}_m$}{$m$}
            \For{each group $g$ of $G$ elements}
                \State $s_g \gets \max(|m_g|)$
                \State $m'_g \gets m_g / s_g$
                \State $m''_g \gets 2 m'_g / (1 + |m'_g|)$
                \State $m^q_g \gets \mathrm{Round}(m''_g \cdot 127, \text{INT8})$
            \EndFor
            \State \Return $m^q, s$
        \EndFunction
        \vspace{0.25\baselineskip}
        \Function{$\mathcal{Q}_m^{-1}$}{$m^q, s$}
            \For{each group $g$}
                \State $m''_g \gets m^q_g / 127$
                \State $m'_g \gets m''_g / (2 - |m''_g|)$
                \State $m_g \gets m'_g \cdot s_g$
            \EndFor
            \State \Return $m$
        \EndFunction
        \end{algorithmic}
        \end{algorithm}

\begin{algorithm}[t]
\caption{$\mathcal{Q}_v$: Variance Quantization}
\label{alg:quant_var}
\begin{algorithmic}[1]
\Statex \textbf{Constants:} Group size $G = 32$
\vspace{0.25\baselineskip}
\Function{$\mathcal{Q}_v$}{$v$}
    \State $v' \gets \sqrt{v}$
    \For{each group $g$ of $G$ elements}
        \State $s_g \gets \max(v'_g)$
        \State $v^q_g \gets \mathrm{Round}((v'_g / s_g) \cdot 255, \text{UINT8})$
    \EndFor
    \State \Return $v^q, s$
\EndFunction
\vspace{0.25\baselineskip}
\Function{$\mathcal{Q}_v^{-1}$}{$v^q, s$}
    \For{each group $g$}
        \State $v'_g \gets (v^q_g / 255) \cdot s_g$
        \State $v_g \gets (v'_g)^2$
    \EndFor
    \State \Return $v$
\EndFunction
\end{algorithmic}
\end{algorithm}

For both momentum and variance, we partition the tensor into groups of $G=32$ elements and store a separate FP16 scale factor per group, introducing an overhead of $2/G = 1/16$ bytes per parameter.
We store the normalized momentum in signed integers (INT8) and variance in unsigned integers (UINT8) since variance is non-negative.
Algorithms~\ref{alg:quant_mom} and~\ref{alg:quant_var} detail the quantization and dequantization procedures for momentum and variance, respectively.

\begin{algorithm}[t]
    \caption{FlashAdamW: Memory-Efficient AdamW. We \hltext{highlight} the changes from AdamW.}
    \label{alg:flashadamw}
    \begin{algorithmic}[1]
    \Require Parameters $\theta_0$, learning rate schedule $\{\eta_t\}_{t=1}^{T}$, $\beta_1, \beta_2 \in [0,1)$, $\varepsilon > 0$, weight decay $\lambda \geq 0$, loss $\mathcal{L}(\theta)$, minibatch sampler $\mathcal{B}(\cdot)$
    \State $\hlchange{m_0^q, m_0^s \gets \mathcal{Q}_m(0)}$
    \State $\hlchange{v_0^q, v_0^s \gets \mathcal{Q}_v(0)}$
    \State $\hlchange{\thetalp_0, \rho_0 \gets \mathcal{C}(\theta_0)}$
    \State Initialize $t \gets 0$
    \While{not converged}
        \State $t \gets t + 1$
        \State $B_t \sim \mathcal{B}$
        \State $g_t \gets \nabla_\theta \mathcal{L}(B_t; \hlchange{\thetalp_{t-1}})$
        \State \Comment{Reconstruct optimizer state and master weight}
        \State $\hlchange{m_{t-1} \gets \mathcal{Q}_m^{-1}(m^q_{t-1}, m^s_{t-1})}$
        \State $\hlchange{v_{t-1} \gets \mathcal{Q}_v^{-1}(v^q_{t-1}, v^s_{t-1})}$
        \State $\hlchange{\theta_{t-1} \gets \mathcal{C}^{-1}(\thetalp_{t-1}, \rho_{t-1})}$
        \State \Comment{Standard optimizer update}
        \State $m_t \gets \beta_1 m_{t-1} + (1 - \beta_1) g_t$
        \State $v_t \gets \beta_2 v_{t-1} + (1 - \beta_2) g_t^2$
        
        \State $\hat{m}_t \gets m_t / (1 - \beta_1^t)$
        \State $\hat{v}_t \gets v_t / (1 - \beta_2^t)$
        
        \State $\theta_t \gets \theta_{t-1} - \eta_t \left( \hat{m}_t / (\sqrt{\hat{v}_t} + \varepsilon) + \lambda \theta_{t-1} \right)$
        \State \Comment{Quantize optimizer state and split master weight}
        \State $\hlchange{m^q_t, m^s_t \gets \mathcal{Q}_m(m_t)}$
        \State $\hlchange{v^q_t, v^s_t \gets \mathcal{Q}_v(v_t)}$
        \State $\hlchange{\thetalp_t, \rho_t \gets \mathcal{C}(\theta_t)}$
    \EndWhile
    \end{algorithmic}
    \end{algorithm}

\subsection{Optimizer Update}
\label{sub:optimizer_update}

We modify any given gradient update rule by adding a prologue and an epilogue. In the prologue, we dequantize the optimizer states and reconstruct the master weight $\theta$ from the low-precision weight $\thetalp$ and the error correction bits $\rho$. In the epilogue, we quantize the new optimizer state and split the new $\theta$ into an updated $(\thetalp, \rho)$. At the start of training, 
we downcast the master weights to BF16
to ensure training runs directly on the low-precision $\thetalp$ with no downcasts apart from our optimizer step.
\autoref{alg:flashadamw} illustrates these changes for the AdamW optimizer, and Algorithms~\ref{alg:flashsgd} and~\ref{alg:flashlion} in the appendix show the corresponding changes to SGD and Lion respectively.

\subsection{Implementation}

\textbf{Update kernels.}
Since compression and quantization are bandwidth-bound operations, we implement the optimizer step as a single fused Triton kernel~\citep{tillet2019triton}. 
For example, for the AdamW update, our kernel encompasses steps 9 through 22 from~\autoref{alg:flashadamw}.

\textbf{Gradient release.}
We implement gradient release~\citep{zhang2023adam}, interleaving gradient computation with optimizer updates during backpropagation.
As each gradient is computed, we eagerly apply the optimizer rule to free the gradient memory.
We apply this optimization only when gradient accumulation is disabled. 

\textbf{Distributed training.}
Our implementation is compatible with parameter sharding approaches such as PyTorch FSDP~\citep{zhao2023pytorchfsdp}.
During forward and backward passes, only the 16-bit $\thetalp$ parameters are all-gathered; the correction term $\rho$ remains local with the optimizer states.

\textbf{Checkpoint size.}
Our representation reduces checkpoint size.
For instance, standard Adam checkpoints require 12 bytes per parameter (4 for weights, 4 for momentum, 4 for variance); \flashAdam requires only 5 bytes (2 for weights, 1 for correction, 1 for momentum, and 1 for variance).
For a 7B model, checkpoint size reduces from 84\,GiB to 35\,GiB.

\textbf{Code availability.}
We make our implementation widely available as an open-source PyTorch library at \url{https://github.com/databricks/flashoptim}.

\section{Experiments} \label{sec:results}

\subsection{Experimental Setup}

We evaluate \flashOptim with three optimizers: SGD with momentum~\citep{polyak1964some}, AdamW~\citep{loshchilov2017decoupled}, and Lion~\citep{chen2023lion}. We refer to these variants as \flashSGD, \flashAdam, and \flashLion. We test these across several large-scale deep learning tasks, including image classification, language model pretraining, and supervised finetuning.

To ensure a fair comparison, all experiments use identical hyperparameters between reference optimizers and their \flashOptim counterparts. We re-implement the reference optimizers with similar fused Triton kernels for consistent measurement of memory and throughput, and all reference implementations use mixed precision~\citep{micikevicius2018mixed} to keep activations in 16-bit precision. 
The precisions of master weights~$\theta$,
    error correction term~$\rho$, gradients~$g$, momentum~$m$,
    variance~$v$, and activations~$a$ are as follows:

\begin{figure}[!h]
    \centering
    \resizebox{\columnwidth}{!}{%
    \begin{tabular}{lcccccc}
        & $\theta$ & $\rho$ & $g$ & $m$ & $v$ & $a$ \\
        \toprule
        Reference     & FP32 & ---  & FP32 & FP32 & FP32  & BF16 \\
        \flashOptim   & BF16 & INT8 & BF16 & INT8 & UINT8 & BF16 \\
        \bottomrule
    \end{tabular}
    }
\end{figure}

Our experiments demonstrate four main findings.
First, \flashOptim matches reference convergence and accuracy across all tested configurations (\S\ref{sec:convergence}).
Second, it reduces optimizer memory by over 50\% with negligible computational overhead (\S\ref{sec:memory}).
Third, our ULP-based weight splitting achieves near-optimal reconstruction (\S\ref{sec:weight_error}).
Finally, our companding functions significantly reduce quantization error for both momentum and variance states (\S\ref{sec:opt_quant}).

\textbf{Image Classification.} We train a ResNet-50 architecture~\citep{he2016resnet} on the ILSVRC2012 (ImageNet-1K) dataset~\citep{deng2009imagenet}. We use the hyperparameters recommended by Nvidia~\citep{nvidia2023resnet}, with additional details provided in Appendix~\ref{sec:appendix_vision}.

\textbf{LLM Pretraining.} We evaluate LLM pretraining using the training recipe outlined in the nanoGPT repository~\citep{karpathy2023nanogpt}.
We use the GPT-2~\citep{radford2019language} architecture and train on a 10B token subset of the FineWeb dataset~\citep{penedo2024fineweb}, following the setup of \citet{jordan2024moddednanogpt}.
We provide hyperparameter details in Appendix~\ref{sec:appendix_llm}.
We evaluate models using a suite of in-context learning (ICL) benchmarks that assess commonsense reasoning and language understanding capabilities. We provide a complete list of benchmarks in Appendix~\ref{sec:appendix_icl}.

\textbf{LLM Finetuning.} We run supervised finetuning on a pretrained Llama-3.1-8B model~\citep{dubey2024llama3} on OpenMathInstruct-2~\citep{toshniwal2024openmathinstruct2}, 
and evaluate on the GSM8k~\citep{cobbe2021gsm8k} benchmark.
We provide further hyperparameter details in Appendix~\ref{sec:appendix_finetuning}.

\textbf{Training and Infrastructure.} We train with distributed data parallelism for the image classification and LLM pretraining tasks, and for LLM finetuning we use FSDP~\citep{zhao2023pytorchfsdp} and activation checkpointing. We train all models using PyTorch 2.8 and CUDA 12.8 on NVIDIA H100 GPUs.
We report the mean and standard deviation for all our results with \nseed random seeds. For loss curve comparisons, we use identical data ordering across methods. System metrics (memory, timing) are measured in steady-state.

\begin{figure}[t]
    \centering

    \begin{subfigure}[t]{\columnwidth}
        \includegraphics[width=\textwidth]{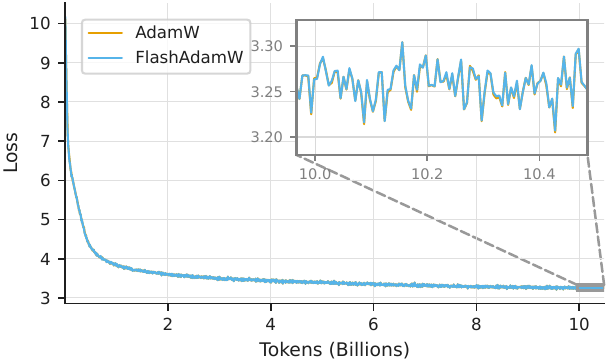}
        \caption{LLM Pretraining (GPT-2 + AdamW)}
        \label{fig:convergence_llm}
    \end{subfigure}
    \begin{subfigure}[t]{\columnwidth}
        \includegraphics[width=\textwidth]{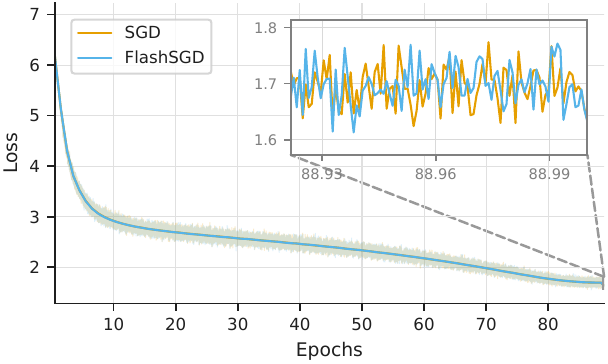}
        \caption{Image Classification (ResNet-50 + SGD)}
        \label{fig:convergence_vision}
    \end{subfigure}
    \caption{
        \textbf{Training convergence}. 
        Comparison of training loss trajectories between reference optimizers and their \flashOptim variants. Both achieve nearly identical loss curves throughout training, demonstrating that our memory optimizations do not impact model quality.}
    \label{fig:convergence}
\end{figure}

\begin{table}[t]
    \centering
    \caption{
    \textbf{Image Classification \& LLM Finetuning Results}.
    Validation accuracy for ResNet-50 (first two columns), and GSM8k accuracy for the LLM finetuning task (last column). We report standard deviation across \nseed training runs. In all settings, \flashOptim matches the reference scores.
    }
    \label{tab:imagenet}
    \small
    \resizebox{\columnwidth}{!}{%
    
\begin{tabular}{lccc}
    & \multicolumn{2}{c}{\textbf{ImageNet Top-1 Acc.}} & \textbf{GSM8k Acc.} \\
    \cmidrule(lr){2-3} \cmidrule(lr){4-4}
    & \textbf{SGD} & \textbf{AdamW} & \textbf{AdamW} \\
    \toprule
    Reference    & 77.01 {\scriptsize $\pm$ 0.02} & 75.51 {\scriptsize $\pm$ 0.09} & 75.09 {\scriptsize $\pm$ 0.40} \\
    \flashOptim  & 77.16 {\scriptsize $\pm$ 0.09} & 75.67 {\scriptsize $\pm$ 0.04} & 74.98 {\scriptsize $\pm$ 0.77} \\
    \bottomrule
\end{tabular}

    }
\end{table}

\begin{table*}
    \centering
    \caption{
    \textbf{LLM Pretraining Results}.
    NanoGPT results with GPT-2 (124M). We report validation loss and accuracy (\%) on in-context learning benchmarks assessing commonsense reasoning and language understanding.
    We report standard deviation across \nseed training runs.
    }
    \label{tab:nanogpt}
    \small
    \resizebox{\textwidth}{!}{%
    
\begin{tabular}{lccccccccc}
    \toprule
    \textbf{Optimizer} & \textbf{Val Loss} & \textbf{HellaSwag} & \textbf{ARC-E} & \textbf{CSQA} & \textbf{PIQA} & \textbf{LAMBADA} & \textbf{Winograd} & \textbf{BoolQ} & \textbf{Mean ICL} \\
    \midrule
    AdamW        & 3.263 {\scriptsize $\pm$ 0.001} & 31.9 {\scriptsize $\pm$ 0.2} & 39.6 {\scriptsize $\pm$ 0.7} & 25.9 {\scriptsize $\pm$ 4.1} & 64.3 {\scriptsize $\pm$ 0.0} & 31.0 {\scriptsize $\pm$ 1.2} & 57.3 {\scriptsize $\pm$ 0.6} & 58.1 {\scriptsize $\pm$ 3.6} & \textbf{44.0 {\scriptsize $\pm$ 0.4}} \\
    \flashAdam   & 3.265 {\scriptsize $\pm$ 0.001} & 31.9 {\scriptsize $\pm$ 0.5} & 39.5 {\scriptsize $\pm$ 0.9} & 30.8 {\scriptsize $\pm$ 2.1} & 64.5 {\scriptsize $\pm$ 0.3} & 31.9 {\scriptsize $\pm$ 0.7} & 59.1 {\scriptsize $\pm$ 1.1} & 57.2 {\scriptsize $\pm$ 4.8} & \textbf{45.0 {\scriptsize $\pm$ 1.0}} \\
    \midrule
    Lion         & 3.240 {\scriptsize $\pm$ 0.002} & 32.3 {\scriptsize $\pm$ 0.0} & 40.0 {\scriptsize $\pm$ 0.5} & 23.3 {\scriptsize $\pm$ 1.8} & 63.8 {\scriptsize $\pm$ 1.0} & 31.5 {\scriptsize $\pm$ 0.2} & 58.9 {\scriptsize $\pm$ 2.0} & 58.1 {\scriptsize $\pm$ 2.4} & \textbf{44.0 {\scriptsize $\pm$ 0.5}} \\
    \flashLion   & 3.240 {\scriptsize $\pm$ 0.001} & 32.4 {\scriptsize $\pm$ 0.3} & 40.8 {\scriptsize $\pm$ 0.2} & 25.4 {\scriptsize $\pm$ 2.9} & 64.2 {\scriptsize $\pm$ 0.5} & 31.6 {\scriptsize $\pm$ 0.5} & 59.1 {\scriptsize $\pm$ 2.4} & 59.1 {\scriptsize $\pm$ 2.3} & \textbf{44.7 {\scriptsize $\pm$ 0.5}} \\
    \bottomrule
\end{tabular}

    }
\end{table*}

\subsection{Convergence and Accuracy} \label{sec:convergence}

We first verify that \flashOptim introduces no measurable degradation by comparing training convergence and validation accuracy.
\autoref{fig:convergence_llm} shows training loss for LLM pretraining with AdamW and \flashAdam.
\flashAdam produces a nearly identical trajectory to the reference AdamW and closely tracks AdamW even after 20,000 parameter updates, indicating that reduced precision does not affect learning dynamics.
\autoref{fig:convergence_vision} shows similar results for image classification: \flashSGD matches reference SGD throughout training.
For LLM finetuning, \autoref{fig:convergence_finetuning} in the appendix shows an analogous result for AdamW.

\autoref{tab:imagenet} reports final scores for the image classification and LLM finetuning tasks. \flashSGD and \flashAdam match the reference optimizer scores in all three settings. \autoref{tab:nanogpt} compares final validation loss and in-context learning scores for the LLM pretraining task. 
Models trained with \flashOptim achieve scores within variance of reference optimizers across all metrics.

\subsection{Memory and Speed} \label{sec:memory}

We compare memory requirements and optimizer step time, demonstrating that \flashOptim reduces peak memory~\citep{pytorch_cuda_memory_stats} without practical overhead. We focus on parameter-related memory (weights, optimizer state, gradients) since activation memory is identical across both settings.
To isolate contributions, we ablate: weight splitting (Weight Split) with full-precision optimizer states, and optimizer state quantization with companding (Opt.\ Quant.) with FP32 master weights.

\autoref{tab:finetuning_profiling} breaks down the memory usage for LLM finetuning on a Llama-3.1-8B model. As anticipated, we reduce parameter memory by 50\% from dropping precision from FP32 to BF16, and reduce optimizer memory by 60\% from quantizing the optimizer state tensors. Moreover, when looking at peak memory (including activations), \flashOptim reduces it by 36\% with no practical slowdown in optimizer step times.

Ablating each component confirms that weight splitting halves master weight memory while adding 12\% of extra optimizer state. Optimizer state quantization reduces optimizer state by $\sim$73\%, mapping FP32 tensors to INT8/UINT8; the reduction is slightly less than 75\% due to the overhead of storing FP16 scale factors for each group of 32 elements.
\autoref{tab:imagenet_profiling} and \autoref{tab:nanogpt_profiling} in the appendix show similar trends for LLM pretraining and image classification.

\begin{table}
    \centering
    \caption{\textbf{Profiling}.
    Runtime measurements for LLM Finetuning with Llama-3.1-8B. We capture master weight memory (Params), optimizer state memory (Optim), peak GPU memory (Peak), and optimizer step times (Step). 
    }
    \label{tab:finetuning_profiling}
    
\footnotesize
\setlength{\tabcolsep}{3pt}
\begin{tabularx}{\columnwidth}{l *{3}{Yr}c}
    \toprule
    & \multicolumn{2}{c}{\textbf{Params}} & \multicolumn{2}{c}{\textbf{Optim}} & \multicolumn{2}{c}{\textbf{Peak}} & \textbf{Step} \\
    \cmidrule(lr){2-3} \cmidrule(lr){4-5} \cmidrule(lr){6-7} \cmidrule(lr){8-8}
    \textbf{Variant} & GiB & $\Delta$ & GiB & $\Delta$ & GiB & $\Delta$ & ms \\
    \midrule
    Reference    & 29.9 &  & 59.8 &  & 175.2 &  & 12.5 \\
    FlashOptim   & \textbf{15.0} & {\color{black!70}\scriptsize -50\%} & \textbf{23.4} & {\color{black!70}\scriptsize -61\%} & \textbf{112.9} & {\color{black!70}\scriptsize -36\%} & 11.5 \\
    \midrule
    \multicolumn{8}{l}{\emph{Ablations}} \\
    Weight Split & 15.0 & {\color{black!70}\scriptsize -50\%} & 67.3 & {\color{black!70}\scriptsize +12\%} & 156.7 & {\color{black!70}\scriptsize -11\%} & 10.7 \\
    Opt. Quant.  & 29.9 &  & 15.9 & {\color{black!70}\scriptsize -73\%} & 131.4 & {\color{black!70}\scriptsize -25\%} & 10.5 \\
    \bottomrule
\end{tabularx}

\end{table}

\subsection{Weight Error Correction} \label{sec:weight_error}

We compare our weight splitting scheme to~\citet{zamirai2020revisiting}, who store the rounding error in a floating-point buffer for Kahan summation error correction. Since both approaches are data-independent, we evaluate them exhaustively over all finite FP32 bitstrings, computing relative error after applying compression and decompression in sequence.
We consider four methods: a no error correction baseline, storing error in the same 16-bit format, our ULP-normalized error with 8-bit integers, and ours with 16-bit integers.

\begin{figure}[t]
    \centering
    \includegraphics[width=\columnwidth]{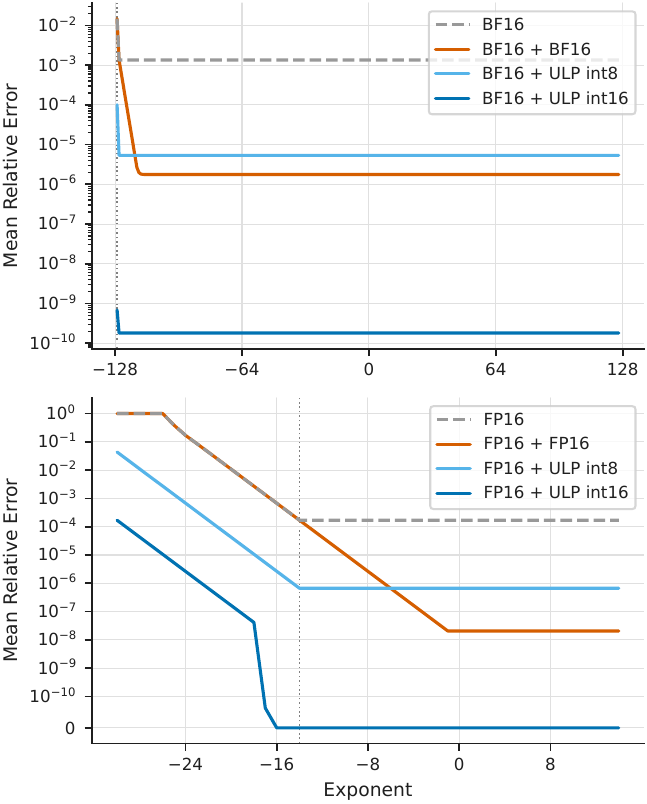}
    \caption{
        \textbf{FP32 Reconstruction Error.}
        Comparison of FP32 reconstruction error for different weight compression schemes across exponent ranges for a target datatype of BF16 (top) and FP16 (bottom).
    Our ULP-based error correction achieves lower relative error, particularly for small exponents. 
    Denormal floating-point ranges are indicated with vertical dotted lines.
    }
    \label{fig:compression_comparison}
\end{figure}

\autoref{fig:compression_comparison} plots mean relative error versus exponent for BF16 and FP16.
For BF16, our ULP approach with 16-bit correction achieves near-zero error ($<10^{-9}$).
With 16 bits of correction term, we achieve perfect bitwise reconstruction in $99.92\%$ of the values.
In contrast, storing the error in BF16 (BF16+BF16) produces substantially worse error ($>10^{-6}$), comparable to our 24-bit format.
For FP16, our 32-bit ULP format perfectly reconstructs values in the normal range and dominates FP16+FP16 throughout.
Our 24-bit format produces constant error across the normal range, improving worst-case error from $10^{-4}$ to under $10^{-6}$.

\subsection{Optimizer State Quantization} \label{sec:opt_quant}

\begin{figure}[t]
    \centering
    \includegraphics[width=\columnwidth]{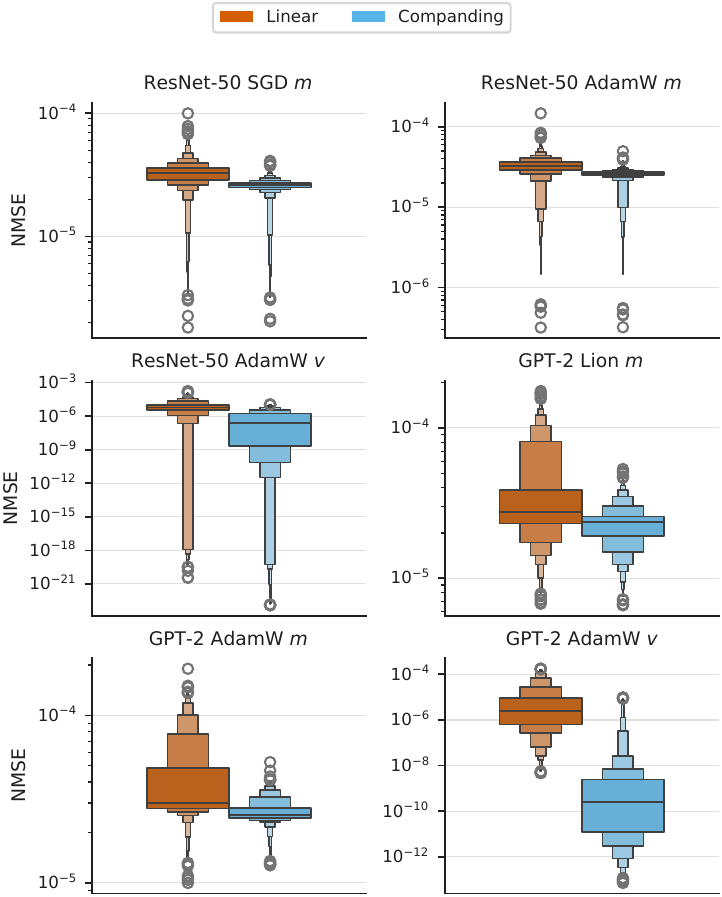}
    \caption{
        \textbf{Optimizer state quantization error.}
        NMSE comparison between standard scaled integer quantization (Linear) and our companding approach for momentum ($m$) and variance ($v$) buffers across different optimizers and datasets.
        Companding reduces quantization error across all optimizer types and tensor types, with particularly large improvements for variance tensors.
    }
    \label{fig:quant_nmse}
\end{figure}

We validate our companding functions by comparing quantization error against standard scaled integer quantization.
Using a fixed full-precision training trajectory for consistency, we quantize and dequantize momentum and variance buffers at each step, computing normalized MSE (NMSE) against the original values. 
\autoref{fig:quant_nmse} shows quantile distributions of NMSE for each optimizer and buffer type.
Companding reduces error for momentum buffers and provides substantial improvements for variance buffers, where NMSE drops significantly.

\begin{figure}[t]
    \centering
    \includegraphics[width=\columnwidth]{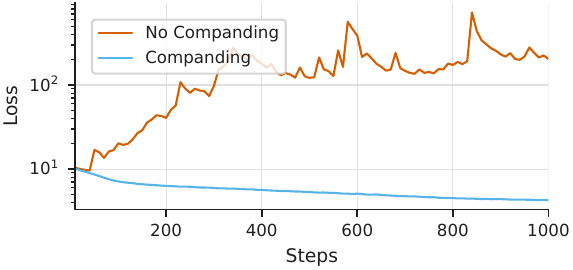}
    \caption{
        \textbf{Companding prevents training divergence.}
        GPT-2 training with AdamW and quantized optimizer states: linear quantization (no companding) causes rapid divergence, while our companding approach maintains stable training dynamics.
    }
    \label{fig:companding_divergence}
\end{figure}

Beyond reducing quantization error, in some cases companding is essential for training stability.
\autoref{fig:companding_divergence} shows LLM pretraining with and without variance companding: linear quantization causes training to diverge, while companding maintains stable convergence.

\section{Limitations} \label{sec:limitations}

\flashOptim is designed to minimize parameter-associated memory consumption, 
so models with large parameter counts and small activations benefit most from 
our optimizations.
Smaller architectures with large activations, such as convolutional networks 
with high-resolution feature maps, are often dominated by activation memory.
In these activation-dominated regimes, a 50\% reduction in parameter memory 
translates to modest total memory savings, and techniques like activation 
checkpointing are more effective for such workloads.

Another limitation is that some tasks and architectures may be more sensitive 
to quantization than those in our benchmarks. The effectiveness of a 
quantization pipeline depends on the data distribution, and there is no guarantee that our method (or any quantization approach) will preserve model quality in all cases. Consequently, our implementation allows selectively disabling compression or excluding specific layers as needed.

Finally, while we find that 24-bit master weights are sufficient in our 
experiments, not even 32 bits are guaranteed to suffice in all cases. Successive (normal) floating-point values differ by a factor of roughly $2^{-\text{num\_mantissa\_bits}}$, and if the ratio of weight update to weight magnitude falls below this threshold, the update will be discarded.

\section{Conclusion} \label{sec:conclusion}

We introduced \flashOptim, a method to reduce the memory footprint of neural 
network training while preserving optimizer semantics and model quality.
\flashOptim provides drop-in replacements for common optimizers and requires 
no additional tuning.

Our approach combines two key techniques. First, we reduce master weights from 32 to 24 bits via improved floating-point error correction. Second, we use companding functions to enable 8-bit optimizer state quantization. Together with 16-bit gradients, these reduce per-parameter memory by over 50\% for AdamW.

We validated \flashOptim on image and language benchmarks using SGD, AdamW, and Lion.
Across all settings, our method matches reference implementations in both loss and accuracy while providing significant memory savings.
\flashOptim composes with FSDP and activation checkpointing, enabling multiplicative benefits for large-scale training.
By lowering memory requirements, \flashOptim enables practitioners and researchers with limited hardware to train larger models than previously feasible.

\section*{Acknowledgements}

We are grateful to Jonathan Frankle, Xing Chen, and Matei Zaharia for their continued support and guidance. We also thank Jialu Liu and Erich Elsen for insightful discussions.

{\small
\bibliographystyle{plainnat}
\bibliography{references}
}

\clearpage
\appendix

\section{Method Details}
\label{sec:appendix_method}

This section provides detailed pseudocode for the \flashOptim version of SGD (\autoref{alg:flashsgd}) and Lion (\autoref{alg:flashlion}).

\begin{algorithm}[!h]
    \caption{FlashSGD: Memory-Efficient SGD. We \hltext{highlight} the changes from SGD.}
    \label{alg:flashsgd}
    \begin{algorithmic}[1]
    \Require Parameters $\theta_0$, learning rate schedule $\{\eta_t\}_{t=1}^{T}$, momentum $\mu \in [0,1)$, weight decay $\lambda \geq 0$, loss $\mathcal{L}(\theta)$, minibatch sampler $\mathcal{B}(\cdot)$
    \State $\hlchange{m_0^q, m_0^s \gets \mathcal{Q}_m(0)}$
    \State $\hlchange{\thetalp_0, \rho_0 \gets \mathcal{C}(\theta_0)}$
    \State Initialize $t \gets 0$
    \While{not converged}
        \State $t \gets t + 1$
        \State $B_t \sim \mathcal{B}$
        \State $g_t \gets \nabla_\theta \mathcal{L}(B_t; \hlchange{\thetalp_{t-1}})$
        \State $\hlchange{m_{t-1} \gets \mathcal{Q}_m^{-1}(m^q_{t-1}, m^s_{t-1})}$
        \State $m_t \gets \mu m_{t-1} + g_t$
        \State $\hlchange{m^q_t, m^s_t \gets \mathcal{Q}_m(m_t)}$
        \State $\hlchange{\theta_{t-1} \gets \mathcal{C}^{-1}(\thetalp_{t-1}, \rho_{t-1})}$
        \State $\theta_t \gets \theta_{t-1} - \eta_t (m_t + \lambda \theta_{t-1})$
        \State $\hlchange{\thetalp_t, \rho_t \gets \mathcal{C}(\theta_t)}$
    \EndWhile
    \end{algorithmic}
    \end{algorithm}

\begin{algorithm}[!h]
    \caption{FlashLion: Memory-Efficient Lion. We \hltext{highlight} the changes from Lion.}
    \label{alg:flashlion}
    \begin{algorithmic}[1]
    \Require Parameters $\theta_0$, learning rate schedule $\{\eta_t\}_{t=1}^{T}$, $\beta_1, \beta_2 \in [0,1)$, weight decay $\lambda \geq 0$, loss $\mathcal{L}(\theta)$, minibatch sampler $\mathcal{B}(\cdot)$
    \State $\hlchange{m_0^q, m_0^s \gets \mathcal{Q}_m(0)}$
    \State $\hlchange{\thetalp_0, \rho_0 \gets \mathcal{C}(\theta_0)}$
    \State Initialize $t \gets 0$
    \While{not converged}
        \State $t \gets t + 1$
        \State $B_t \sim \mathcal{B}$
        \State $g_t \gets \nabla_\theta \mathcal{L}(B_t; \hlchange{\thetalp_{t-1}})$
        \State $\hlchange{m_{t-1} \gets \mathcal{Q}_m^{-1}(m^q_{t-1}, m^s_{t-1})}$
        \State $u_t \gets \mathrm{sign}(\beta_1 m_{t-1} + (1 - \beta_1) g_t)$ 
        \State $m_t \gets \beta_2 m_{t-1} + (1 - \beta_2) g_t$
        \State $\hlchange{m^q_t, m^s_t \gets \mathcal{Q}_m(m_t)}$
        \State $\hlchange{\theta_{t-1} \gets \mathcal{C}^{-1}(\thetalp_{t-1}, \rho_{t-1})}$
        \State $\theta_t \gets \theta_{t-1} - \eta_t (u_t + \lambda \theta_{t-1})$
        \State $\hlchange{\thetalp_t, \rho_t \gets \mathcal{C}(\theta_t)}$
    \EndWhile
    \end{algorithmic}
    \end{algorithm}

\section{Experimental Details}

For all our experiments, we use the MosaicML Streaming~\citep{mosaicml2022streaming} library to ensure deterministic data loading for distributed training.

\subsection{Image Classification}
\label{sec:appendix_vision}

We train the ResNet-50~\citep{he2016resnet} model using the \texttt{timm} library on the ILSVRC2012 (ImageNet-1K) dataset~\citep{deng2009imagenet}, which contains approximately 1.28 million training images and 50,000 validation images across 1,000 classes. Our implementation of ImageNet follows the standard setup from~\citep{krizhevsky2012imagenet,simonyan2014very}. The image is resized with its shorter side randomly sampled in $[256, 480]$ for scale augmentation~\citep{simonyan2014very}. A $224 \times 224$ crop is randomly sampled from an image or its horizontal flip, and then normalized. For evaluation, the image is first resized to $256 \times 256$, followed by a $224 \times 224$ center crop, and then normalized.
We initialize the network with Kaiming He initialization~\citep{he2016identity} and zero-init residuals~\citep{he2016resnet}.

We train for 90 epochs with a batch size of 1024, using a 5-epoch linear warmup followed by cosine learning rate decay, following the recommended settings from~\citep{goyal2017accurate}. We disable weight decay for biases and BatchNorm layers. We apply label smoothing~\citep{szegedy2016rethinking} with coefficient 0.1. Both reference (FP32 master weights) and \flashOptim use BF16 activations. \autoref{tab:vision_hparams} summarizes the hyperparameters we use for training.

\begin{table}[ht]
\centering
\caption{Optimizer hyperparameters for ImageNet/ResNet-50.}
\label{tab:vision_hparams}
\small
\begin{tabular}{@{}lcc@{}}
\toprule
& SGD & AdamW \\
\midrule
Learning Rate & 1.024 & $3 \times 10^{-3}$ \\
Momentum / Betas & 0.9 & (0.9, 0.999) \\
Weight Decay & $3 \times 10^{-5}$ & $3 \times 10^{-4}$ \\
\bottomrule
\end{tabular}
\end{table}

\paragraph{Memory and Speed Profiling.}
\autoref{tab:imagenet_profiling} presents the results of the memory and speed profile for training.

\begin{table}[ht]
    \centering
    \caption{Memory and speed profiling for image classification (ResNet-50).}
    \label{tab:imagenet_profiling}
    
\footnotesize
\setlength{\tabcolsep}{3pt}
\begin{tabularx}{\columnwidth}{l *{3}{Yr}c}
    \toprule
    & \multicolumn{2}{c}{\textbf{Params}} & \multicolumn{2}{c}{\textbf{Optim}} & \multicolumn{2}{c}{\textbf{Total}} & \textbf{Step} \\
    \cmidrule(lr){2-3} \cmidrule(lr){4-5} \cmidrule(lr){6-7} \cmidrule(lr){8-8}
    \textbf{Variant} & GiB & $\Delta$ & GiB & $\Delta$ & GiB & $\Delta$ & ms \\
    \midrule
    \multicolumn{8}{l}{\textbf{SGD}} \\
    \quad Reference    & 0.10 &  & 0.10 &  & 0.30 &  & 8.4 \\
    \quad FlashOptim   & 0.05 & {\color{black!70}\scriptsize -46\%} & 0.05 & {\color{black!70}\scriptsize -45\%} & 0.17 & {\color{black!70}\scriptsize -45\%} & 9.0 \\
    \quad Weight Split & 0.05 & {\color{black!70}\scriptsize -46\%} & 0.12 & {\color{black!70}\scriptsize +23\%} & 0.23 & {\color{black!70}\scriptsize -22\%} & 8.7 \\
    \quad Opt. Quant.  & 0.10 &  & 0.03 & {\color{black!70}\scriptsize -73\%} & 0.23 & {\color{black!70}\scriptsize -23\%} & 8.7 \\
    \midrule
    \multicolumn{8}{l}{\textbf{AdamW}} \\
    \quad Reference    & 0.10 &  & 0.19 &  & 0.40 &  & 11.9 \\
    \quad FlashOptim   & 0.05 & {\color{black!70}\scriptsize -46\%} & 0.08 & {\color{black!70}\scriptsize -56\%} & 0.20 & {\color{black!70}\scriptsize -50\%} & 12.2 \\
    \quad Weight Split & 0.05 & {\color{black!70}\scriptsize -46\%} & 0.21 & {\color{black!70}\scriptsize +11\%} & 0.33 & {\color{black!70}\scriptsize -17\%} & 12.1 \\
    \quad Opt. Quant.  & 0.10 &  & 0.05 & {\color{black!70}\scriptsize -73\%} & 0.25 & {\color{black!70}\scriptsize -36\%} & 12.6 \\
    \bottomrule
\end{tabularx}

\end{table}

\begin{figure}[b]
    \centering
    \includegraphics[width=\columnwidth]{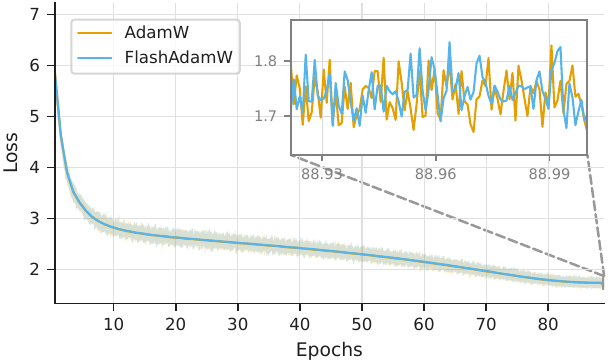}
    \caption{
        \textbf{Training convergence for image classification} (ResNet-50 + AdamW).
        Comparison of training loss between reference AdamW and \flashAdam on ImageNet.
    }
    \label{fig:convergence_vision_adamw}
\end{figure}

\autoref{fig:convergence_vision_adamw} shows training loss for ResNet-50 with AdamW and \flashAdam. \flashAdam produces a nearly identical trajectory to the reference AdamW implementation.

\subsection{LLM Pretraining}
\label{sec:appendix_llm}

We train the GPT-2~\citep{radford2019language} (124M) architecture with 12 transformer layers, 12 attention heads, embedding dimension 768, and context length of 1024. We train on the FineWeb10B~\citep{kjj0_fineweb10b_gpt2} dataset, a subset of approximately 10 billion tokens from the FineWeb~\citep{penedo2024fineweb} dataset, tokenized using the GPT-2 BPE tokenizer. We train for 20,000 steps with a batch size of roughly 0.5 million tokens per step. We apply learning rate warmup for the first 700 steps, followed by a cosine decay to 0. The global norm is clipped at 1.0 and a weight decay of 0.1 is used. Weight decay is applied only to 2D parameters (i.e., weight matrices and embeddings), excluding biases and layer normalization parameters. We train in BF16 mixed precision. \autoref{tab:llm_hparams} summarizes the optimizer hyperparameters.

\begin{table}[ht]
\centering
\caption{Optimizer hyperparameters for GPT-2 124M pretraining.}
\label{tab:llm_hparams}
\small
\begin{tabular}{@{}lcc@{}}
\toprule
& AdamW & Lion \\
\midrule
Learning Rate & $6 \times 10^{-4}$ & $2 \times 10^{-4}$ \\
Betas & (0.9, 0.95) & (0.9, 0.95) \\
\bottomrule
\end{tabular}
\end{table}

\paragraph{Memory and Speed Profiling.}
\autoref{tab:nanogpt_profiling} presents the memory and speed profiling results for LLM pretraining.

\begin{table}[ht]
    \centering
    \caption{Memory and speed profiling for LLM pretraining (GPT-2 124M).}
    \label{tab:nanogpt_profiling}
    
\footnotesize
\setlength{\tabcolsep}{3pt}
\begin{tabularx}{\columnwidth}{l *{3}{Yr}c}
    \toprule
    & \multicolumn{2}{c}{\textbf{Params}} & \multicolumn{2}{c}{\textbf{Optim}} & \multicolumn{2}{c}{\textbf{Total}} & \textbf{Step} \\
    \cmidrule(lr){2-3} \cmidrule(lr){4-5} \cmidrule(lr){6-7} \cmidrule(lr){8-8}
    \textbf{Variant} & GiB & $\Delta$ & GiB & $\Delta$ & GiB & $\Delta$ & ms \\
    \midrule
    \multicolumn{8}{l}{\textbf{AdamW}} \\
    \quad Reference    & 0.46 &  & 0.93 &  & 1.77 &  & 5.7 \\
    \quad FlashOptim   & 0.23 & {\color{black!70}\scriptsize -50\%} & 0.36 & {\color{black!70}\scriptsize -61\%} & 0.74 & {\color{black!70}\scriptsize -58\%} & 5.9 \\
    \quad Weight Split & 0.23 & {\color{black!70}\scriptsize -50\%} & 1.04 & {\color{black!70}\scriptsize +12\%} & 1.43 & {\color{black!70}\scriptsize -20\%} & 5.9 \\
    \quad Opt. Quant.  & 0.46 &  & 0.25 & {\color{black!70}\scriptsize -73\%} & 1.08 & {\color{black!70}\scriptsize -39\%} & 5.8 \\
    \midrule
    \multicolumn{8}{l}{\textbf{Lion}} \\
    \quad Reference    & 0.46 &  & 0.46 &  & 1.30 &  & 4.3 \\
    \quad FlashOptim   & 0.23 & {\color{black!70}\scriptsize -50\%} & 0.24 & {\color{black!70}\scriptsize -48\%} & 0.62 & {\color{black!70}\scriptsize -53\%} & 4.5 \\
    \quad Weight Split & 0.23 & {\color{black!70}\scriptsize -50\%} & 0.58 & {\color{black!70}\scriptsize +25\%} & 0.96 & {\color{black!70}\scriptsize -26\%} & 4.4 \\
    \quad Opt. Quant.  & 0.46 &  & 0.12 & {\color{black!70}\scriptsize -73\%} & 0.96 & {\color{black!70}\scriptsize -26\%} & 4.4 \\
    \bottomrule
\end{tabularx}

\end{table}

\begin{figure}[b]
    \centering
    \includegraphics[width=\columnwidth]{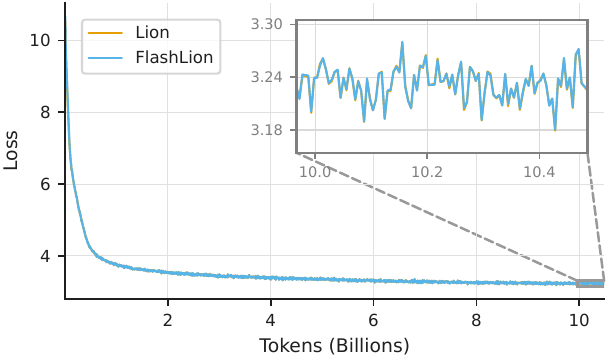}
    \caption{
        \textbf{Training convergence for LLM pretraining} (GPT-2 + Lion).
        Comparison of validation loss between reference Lion and \flashLion on FineWeb10B.
    }
    \label{fig:convergence_llm_lion}
\end{figure}

\autoref{fig:convergence_llm_lion} shows training loss for LLM pretraining with Lion and \flashLion. \flashLion produces a nearly identical trajectory to the reference Lion implementation and closely tracks Lion even after 20,000 parameter updates.

\subsection{In-Context Learning Benchmarks}
\label{sec:appendix_icl}

We evaluate our pretrained language models on a suite of eight in-context learning~\citep{brown2020language} benchmarks that assess diverse commonsense reasoning and language understanding capabilities:

\begin{itemize}[noitemsep]
    \item \textbf{HellaSwag}~\citep{zellers2019hellaswag}: A sentence completion benchmark requiring physical and temporal commonsense.
    \item \textbf{ARC-Easy}~\citep{clark2018arc}: The easy subset of the AI2 Reasoning Challenge, containing grade-school science questions.
    \item \textbf{CommonsenseQA}~\citep{talmor2019commonsenseqa}: Multiple-choice questions requiring commonsense knowledge from ConceptNet.
    \item \textbf{PIQA}~\citep{bisk2020piqa}: Physical Interaction Question Answering, testing physical commonsense reasoning.
    \item \textbf{OpenBookQA}~\citep{mihaylov2018openbookqa}: Elementary science questions requiring multi-step reasoning over facts.
    \item \textbf{LAMBADA}~\citep{paperno2016lambada}: Word prediction requiring broad discourse context understanding.
    \item \textbf{Winograd}~\citep{levesque2012winograd}: Pronoun resolution problems requiring commonsense reasoning.
    \item \textbf{BoolQ}~\citep{clark2019boolq}: Naturally occurring yes/no reading comprehension questions.
\end{itemize}

All benchmarks are evaluated in a zero-shot setting.

\subsection{LLM Finetuning}
\label{sec:appendix_finetuning}

We fine-tune Llama-3.1-8B~\citep{dubey2024llama3} on OpenMathInstruct-2~\citep{toshniwal2024openmathinstruct2}. For evaluation, we use GSM8k~\citep{cobbe2021gsm8k}, a benchmark of 1,319 grade school math word problems that require multi-step arithmetic reasoning.

\paragraph{Training and Evaluation.}
We use FSDP2~\citep{fsdp2} with full parameter sharding and activation checkpointing~\citep{chen2016training} applied to every transformer layer. We use the AdamW optimizer, with $\beta_1=0.9$ and $\beta_2=0.95$. The global norm is clipped at 1.0 and a weight decay of 0.1 is used. Weight decay is applied only to weight matrices, excluding biases, embeddings, and layer normalization parameters. We train for 5000 steps with an effective batch size of approximately 5.2 million tokens per step. We apply learning rate warmup for the first 1000 steps, followed by a cosine decay to 0.

We evaluate on the GSM8k test set using temperature $T=0.2$ decoding.
Following standard practice, we extract the final numerical answer from model generations and compare against ground truth.

\begin{figure}[h]
    \centering
    \includegraphics[width=\columnwidth]{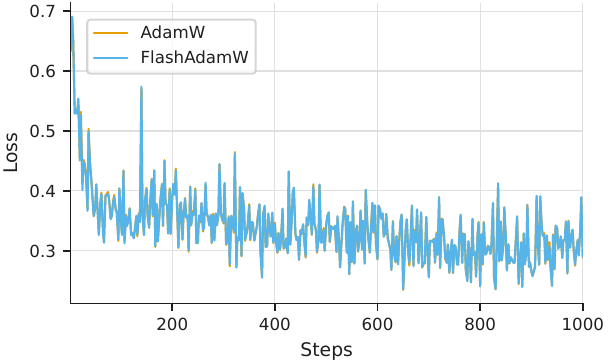}
    \caption{
        \textbf{Training convergence for LLM finetuning} (Llama-3.1-8B + AdamW).
        Comparison of training loss between reference AdamW and \flashAdam during supervised finetuning on OpenMathInstruct-2.
    }
    \label{fig:convergence_finetuning}
\end{figure}

\end{document}